\documentclass[acmlarge]{acmart}

\usepackage{algorithm}
\usepackage{algpseudocode}

\AtBeginDocument{%
  \providecommand\BibTeX{{%
    \normalfont B\kern-0.5em{\scshape i\kern-0.25em b}\kern-0.8em\TeX}}}

\copyrightyear{2024}

\title{An Evaluation of Real-time Adaptive Sampling Change Point Detection Algorithm using KCUSUM}

\begin{document}
\author{Vijayalakshmi Saravanan}
\authornote{Corresponding first author}
\email{vijayalakshmi.sarava@usd.edu}
\affiliation{%
  \institution{University of South Dakota}
  \streetaddress{414E Clark Street}
  \city{Vermillion}
  \country{USA}}

\author{Perry Siehien}
\affiliation{%
   \institution{University of South Dakota}
  \streetaddress{414E Clark Street}
  \city{Vermillion}
  \country{USA}}
\email{perry.siehien@coyotes.usd.edu}
\author{Shinjae Yoo}
\affiliation{%
\institution{Brookhaven National Lab}
 \state{NY}
 \country{USA}}
 \email{sjyoo@bnl.gov}

\author{Hubertus Van Dam}
\affiliation{%
 \institution{Brookhaven National Lab}
 \state{NY}
 \country{USA}}
 \email{hvandam@bnl.gov}

\author{Thomas Flynn}
\affiliation{%
 \institution{Brookhaven National Lab}
 \state{NY}
 \country{USA}}
\email{tflynn@bnl.gov}
\author{Christopher Kelly}
\affiliation{%
 \institution{Brookhaven National Lab}
 \state{NY}
 \country{USA}}
 \email{ckelly@bnl.gov}
\author{Khaled Z Ibrahim}
\affiliation{
 \institution{Lawrence Berkeley National Lab}
 \state{Berkeley}
 \country{USA}}
\email{kzibrahim@lbl.gov}

\begin{abstract}
Detecting abrupt changes in real-time data streams from scientific simulations presents a challenging task, demanding the deployment of accurate and efficient algorithms. Identifying change points in live data stream involves continuous scrutiny of incoming observations for deviations in their statistical characteristics, particularly in high-volume data scenarios. Maintaining a balance between sudden change detection and minimizing false alarms is vital. Many existing algorithms for this purpose rely on known probability distributions, limiting their feasibility. In this study, we introduce the Kernel-based Cumulative Sum (KCUSUM) algorithm, a non-parametric extension of the traditional Cumulative Sum (CUSUM) method, which has gained prominence for its efficacy in online change point detection under less restrictive conditions. KCUSUM splits itself by comparing incoming samples directly with reference samples and computes a statistic grounded in the Maximum Mean Discrepancy (MMD) non-parametric framework. This approach extends KCUSUM's pertinence to scenarios where only reference samples are available, such as atomic trajectories of proteins in vacuum, facilitating the detection of deviations from the reference sample without prior knowledge of the data's underlying distribution. Furthermore, by harnessing MMD's inherent random-walk structure, we can theoretically analyze KCUSUM's performance across various use cases, including metrics like expected delay and mean runtime to false alarms. Finally, we discuss real-world use cases from scientific simulations such as NWChem CODAR and protein folding data, demonstrating KCUSUM's practical effectiveness in online change point detection.
\end{abstract}



\received{XX February 2024}
\received[revised]{XXX}
\received[accepted]{XXX}

\maketitle

\section{Introduction}

The detection of change points on streaming data without distributional information is an ongoing effort, with further difficulty introduced by increasingly larger volumes of data. In our work, we are interested in mitigating the potential issues imposed by these conditions, while still maintaining robust change detection. Detecting change points on live data could mean detecting a change from the average value or variance, or any other statistical property. In particular, the intended application on streaming data requires a change to be detected shortly after it occurs, without knowledge of the entire dataset. Examples of continuous datasets include quality control systems, scientific equipment, or biochemical simulations, among others.

There may be cases where sufficient knowledge of the data is available, for which an optimal algorithm can be chosen for live change detection. For example, if the probability distributions of before and after the change are known, the Cumulative Sum (CUSUM) algorithm (Algorithm II.I) is an optimal procedure, due to its short delay and minimal false alarm rate \cite{Lorden71,Lorden70}. Apart from optimality, CUSUM is a simple algorithm to program and its output of maximum likelihood is easy to interpret. In some cases, the relevant statistical properties cannot be modeled directly from the data, and would be difficult to implement CUSUM without significant modification.

A similar topic to change detection is statistical hypothesis testing. In our work, we leverage established methods for two-sample testing and adjust them for change detection. Generally, two-sample tests are non-parametric evaluations used to determine whether samples are drawn from the same distribution. For the task of online change detection, our approach relies on kernel embedding to calculate the Maximum Mean Discrepancy (MMD) statistic. By using the MMD framework, two datasets can be compared by computing the distance between corresponding observables, calculated by a positive definite kernel, which is introduced in Section \ref{mmd}. The distance found with MMD is an unbiased estimator and is not restricted to Euclidean datasets. This framework can be further generalized to hypothesis testing that involves strings, graphs or other structured datatypes \cite{Muandet07}.

Since previous work has demonstrated the MMD framework in hypothesis testing, we introduce our Kernel Cumulative Sum (KCUSUM) algorithm (Algorithm V.I). In contrast to CUSUM, KCUSUM does not require knowledge of the sample distribution before and after a change occurs. Instead, KCUSUM relies on samples from the pre-change regime and continuously compares them to incoming data using a kernel chosen by the user. By utilizing this framework, KCUSUM can detect changes to any distribution when the distance from the pre-change distribution exceeds a user-defined threshold. Initial theoretical results concern detection delay and average time to false alarm of the proposed algorithm. These results show the derived upper bound for detection time and the expected time until false alarm when no changes are present. The proceeding analysis of CUSUM relies on prior work for change detection \cite{Lorden70}.

The rest of our work is presented as follows. In the next section, we review some basic properties and behaviors of the CUSUM algorithm. In Section \ref{mmd}, the MMD framework is introduced. KCUSUM is reviewed in Section \ref{kcum} where further analysis is presented. In Section \ref{res}, we show the results of using the KCUSUM algorithm on several datasets that cover expected use cases.

\section{Cumulative Sum Algorithm} \label{cum}

Let us observe a sequence of random variables \(X_{n\geq1}\) with index t such that for all 1 \(\leq i < t\), the random variables \(X_i\) have the distribution \(p_o\), and for \(i \geq t\) have the distribution \(p_1\). We assume these variables have values in Euclidean space \(\mathbb{R}^d\), although the proposed kernel variant is not restricted to these parameters. The index at \(t\) is the change point with this convention, which is the detection target when performing analysis with the CUSUM algorithm. The analysis is performed in real time, where CUSUM detects if a change has occurred by time \(n\), on observables for a sequence of length \(n\).

CUSUM (Cumulative Sum algorithm) is a live change point detection method proposed by Page \cite{Page54}. To show how the algorithm detects changes, let us assume two distributions, \(p_o\) and \(p_1\), which have the densities \(f_o\) and \(f_1\). The full procedure of CUSUM is shown in ALGO II.I. For each step \(n\), data (\(X_n\)) is observed, and a log-likelihood ratio \(log\frac{f_o (x)}{f_1(x)}\) is calculated, with each result added to a threshold statistic \(Z_n\). If a negative value is calculated, \(Z_n\) becomes 0, which restarts the running calculation. If \(Z_n\) reaches threshold \(h\), then a change has been detected at time \(n\). The algorithm ends at time \(T_{CUSUM}\):

\centerline{\(T_{CUSUM} = inf(n\geq 1 | Z_n \geq h) \)}

To better see how the algorithm works, consider the calculation of the log likelihood ratio before and after the change point t. Preceding any change in the distribution, it has mean \(-dKL(p_o, p_1)\), where \( dKL(p_o, p_1) = \mathbb{E}_{P_o} log\frac{f_o(x)}{f_1(x)}\) is the Kullback-Liebler divergence between the two distributions. If \(dKL\) is positive when \(p_o  \neq p_1\), the running total will have a negative average. A negative momentum with the barrier at 0 causes the value to remain near 0 before the distribution changes. If the distribution does change, the value has a positive average \(dKL(po, p1)\) and \(Z_n\) will accumulate the statistic, eventually reaching a threshold h with guaranteed probability, ending the algorithm. Additionally, the CUSUM algorithm is optimized for certain tasks, as we review below.

\begin{center}
\begin{minipage}{.6\textwidth}
\begin{algorithm}[H]
    \caption{\textbf{-II.I:} Cumulative Sum (CUSUM)}\label{cusumAlgo}
        \begin{algorithmic}
        \State \textbf{input:} Data \(x_1,x_2,\ldots\) and threshold \(h\geq0\)
        \State \textbf{initialize} \(Z_o=0\)
        \State \textbf{For} \(n = 1,2,\ldots\)\textbf{do}
                \State \hspace{3mm} \(Z_n=\max\{0,Z_{n-1}+log\frac{f_1(x_n)}{f_o(x_n)}\}\)
                \State \hspace{4mm} \textbf{if} \(Z_n\geq h\) \textbf{then}\ set \(T_{CUSUM}=n\) and \textbf{exit}
                \State \hspace{3mm} \textbf{else} continue
                \State \textbf{end} 
        \end{algorithmic}
\end{algorithm}
\end{minipage}
\end{center}

\-\hspace{1cm}\(\textbf{Example\hspace{1mm} II.I}\)

Consider a normal distribution \(N(a,b)\) with mean a and variance \(b\). Let us assume we are detecting a change in the variance of this distribution, with pre-change parameters of \(N(1,1)\) and post-change parameters of \(N(1,4)\). A sample sequence \(X1, . . . , Xn\) of length \(n = 400\) with the change in variance at \(t = 200\) is generated for this example.  We can then calculate the log likelihood, whose value has a negative mean before the change occurs, and a positive mean afterwards \(t = 200\). The threshold for CUSUM, \(Z_n\), gives a detection time at \(T_{CUSUM} = 212\) with \(h = 10\).

In this section, we review the performance characteristics of the CUSUM algorithm. Every change point defines a unique distribution on observables \(\{X_{n}\}_{n\geq1}\). If the distribution doesn't change, the sequences are independent but identically distributed (i.i.d.) with \(X_i ~ p_o\) for all \(i \geq 1\). This distribution is denoted on sequences by \(\mathbb{P}_\infty\), with the respective expectations \(\mathbb{E}_\infty\). Generally, a change occurring at \(t \geq 1\) means that for all \(1 \leq i \leq t -1, Xi\)  is i.i.d. with \(X_i ~ p_o\), and for \(i \geq t\), they are i.i.d. with \(X_i ~ p_1\). Assume that \(\mathbb{P}_t\) is the probability distribution on a sequence with a known change at time \(t\), where \(\mathbb{E}_t\) represents the expectation value for the sequence. For \(n\geq1\), let \(F_n\) be the \(\sigma\)-algebra \(F_n = \sigma(X_1,..,X_n)\). \(F_n\) represents any information for our observations up to \(n\). Then, the time of change detection \(T\) can be shown with by the filtration \(\{F_n\}_{n\geq1}\).

When using a change point detector on any given observables, there are two potential errors that can occur. These errors include a false alarm, where the change is detected prematurely, and a delay, where the change is detected after it has already occurred. By utilizing the metrics of average run length to false alarm (ARL2FA) and worst case detection delay (ESADD), we are able to formalize the levels of false alarm and delay.\cite{Moustakides09}.

Given any change detection algorithm T, the expected ARL2FA is:

\begin{equation}
    ARL2FA = \mathbb{E}_\infty [T]
\end{equation}

ARL2FA is the mean detection time given observables without change. One method to measure this delay is utilizing Lorden’s criterion \cite{Lorden71}. Assume that at time \(t \geq 1\), a change occurs. The expected delay with regards to length of the observed sequence up to \(t - 1\) has the value \(\mathbb{E}_t [(T-t)^+ |F_{t-1}]^1\). We can show the worst case delay for any change at time \(t\) by calculating the essential supremum over all sequences of length \(t-1\), given by esssup \(\mathbb{E}_t [(T-t)+|F_{t-1}]\). We can now calculate ESADD:

\begin{equation}
    ESADD = sup_{1 \leq t < \infty} \-\ esssup\-\ \mathbb{E}_t [(T-t)^+ | \mathcal{F}_{t-1}]
\end{equation}

In particular, the CUSUM algorithm optimizes the relationship between ARL2FA and ESADD. Further derivation of the proof can be found in \cite{Lorden71}. The relationship between thresholds \(h\) and ARL2FA and ESADD is non-trivial and requires the determination of complex integration \cite{Page54,Moustakides86}. Nevertheless, there are practical upper and lower bounds that can be derived if given a sequence of length \(n\). To formalize the analysis for future comparison to KCUSUM, let's determine some quantitative boundaries for the performance of CUSUM.

\-\hspace{1cm}\(\textbf{Proposition\hspace{1mm} II.II}\)

The performance of the CUSUM algorithm can be shown to have the following bounds. If the average time to false alarm satisfies the inequality \(ARL2FA_{CUSUM}\geq e^h\), then it can also be shown that  \(\mathbb{E}_1 [((log\frac{f_o (x)}{f_1(x)})^+)^2] < \infty\). Therefore, the following inequality is also true:

\begin{equation}
    ESADD_{CUSUM}\leq \frac{h}{d_{KL}(p_1,p_o)} + \frac{h}{d_{KL}(p_1,p_o)^2} \mathbb{E}_1 [((log\frac{f_1 (x)}{f_o(x)})^+)^2]
\end{equation}

We can see that by raising the user defined threshold \(h\), we increase the time until a false alarm at the cost of a longer detection delay. Accordingly, one observation of the ESADD formula to note is the behavior of detection delays. As two samples' distributions approach similar parameters, the delay will increase. Later in this work, we present similar bounds for the KCUSUM algorithm. In the following section, we discuss the Maximum Mean Discrepancy (MMD) statistic before introducing the kernel variant CUSUM algorithm.

\section{Maximum Mean Discrepancy} \label{mmd}
In statistics, a two-sample test is used to determine whether two datasets are drawn from the same distribution. If enough data is available, the we can assume empirical measurements approximate the true distance within the distribution. As this distance becomes grows, we become more confident that the datasets come from unique distributions. Many classical statistical testing methods, such as the Kolmogorov-Darling \cite{Feller48}, Cramer-von-Mises \cite{Anderson62} and the Anderson-Darling \cite{Pettitt76} tests, rely on this principle.

Similarly, our proposed MMD approach is also based on computing empirical distances between distributions. With MMD, the data is implicitly embedded in a Reproducing Kernel Hilbert Space (RKHS), that corresponds to a user-defined kernel function \(k\) where the embedding distance is calculated \cite{Gretton12}. Compared to classical testing methods, MMD leverages several features that makes it appealing for non-parametric evaluations. First, the computed MMD value has a range of unbiased estimators (SEE DEFINTION OF \(P_L\)). Secondly, additional versatility is offered by which kernel the user choses, which can applied to data without a natural Euclidean form, such as graphs, strings or other structured datatypes \cite{Gartner03,Vichy10}.

Assume that \(\chi\) is a set, such that \(k: \chi \times \chi \rightarrow \mathbb{R}\) is a kernel on this set. This kernel is a symmetric, positive definite function, that can be regarded as a measure of similarity. One may consider the set \(\chi = \mathbb{R}^n\) and the Gaussian kernel:

\begin{equation}
    k(x,y) = e^{\frac{-||x-y||^2}{2}}
\end{equation}

Other choices for the kernel function are shown in \cite{Gretton12}. Let us assume that \(\chi\) has the structure of measurable space \((\chi,\Sigma)\) and that the chosen kernel \(k\) is a measurable function on \(\chi \times \chi\) with the \(\theta\)-algebra product. We define \(P(\chi)\) to be the set of all probability measures on \((\chi,\Sigma)\), and by using \(k\), the subset \(P_k = \{\mu \in P(\chi) | \mathbb{E}_{\chi~\mu}[\sqrt{k(x,x)}] < \infty \) is defined. If the chosen kernel is characteristic, then we can define the MMD statistic on \(P_k (\chi)\), given as \(d_k\). This value is given by the formula:

\begin{equation}
    d_k(p_o,p_1) = \sqrt{\mathbb{E}_{p_oxp_o}[k(x,x`)] +\mathbb{E}_{p_1xp_1}[k(y,y`)] -2\mathbb{E}_{p_oxp_1}[k(x,y)]}
\end{equation}

\centerline{See \cite{Gretton12} for complete details}

One of the unbiased estimators of \(d^2k\) shown in \cite{Gretton12} is the linear statistic \(p_L\) defined below. For convenience, the estimator is defined as the following function \(g\):

\begin{equation}
    g((x_o,x_1),(y_o,y_1) = k(x_o,x_1) + k(y_o,y_1) -k(x_o,y_1) - k(x_1,y_o)
\end{equation}

Now, consider two datasets \(X = \{x_1,\ldots,x_n\} \subseteq \chi\) and \(Y = \{y_1,\ldots,y_n\} \subseteq \chi \). Then, \(P_L(X,Y)\) becomes:

\begin{equation}
    P_L(X,Y) = \frac{1}{n\slash 2}\sum_{i=1}^{n\slash 2}g((x_{2i-1},x_{2i}),(y_{2i-1},y_{2i}))
\end{equation}

\section{Related Work}
The topic of live change point detection was introduced as a method to monitor quality management systems \cite{Shewhart31} and sequential hypothesis testing \cite{Wald47}. In contrast to online change detection methods, offline detection procedures must wait until the entire sequence is observed, and all data becomes available to make a decision. One method of offline change point analysis was explored for information processing systems \cite{Harchaoui09}, where every possible  change time is used to partition the samples into two groups. The two samples consist of pre and post-change observations, where they are compared using various kernels. This process is performed across every partition, which can detect discrepancies between the sample parameters. Then formally, a change is detected at the largest discrepancy. This process only relies on the stream of observables and their direct comparison.

Several approaches to live change detection were suggested in previous research topics \cite{Brodsky93, Li15}. One limitation of this work is that it can only find changes in the mean of the distribution. In many cases, we wish to detect changes in variance or mean, which can be implemented by combining the principles of MMD, with generalized control charts \cite{Shewhart31}. This method utilizes a moving window of predetermined size, similar to convolutions.

Additionally, current research \cite{Polunchenko10} aims to deploy AI-based change detectors on streaming data. It is argued that the CUSUM algorithm and its generalizations are simple neural networks, which can be leveraged with machine learning procedures for more robust change detection. By utilizing the CUSUM algorithm, this method attaches itself to rely on prior knowledge of the observed distribution. This limits practicality in our target use cases, where distributional information may be unavailable.

Many change detection algorithms exist with various optimal behaviors depending on target use cases. While Cumulative Sum aims to reduce ESADD, Shiryaev-Roberts minimizes average delay \cite{Haug22}. Since both algorithms have a recursive form, it may be possible to extend our research to Shiryaev-Roberts change detectors as well, but the topic is still being investigated.

\section{Kernel Cumulative Sum Algorithm} \label{kcum}
The kernel variant of the Cumulative Sum algorithm (KCUSUM) combines properties of the CUSUM algorithm with the proposed MMD procedure. Instead of taking each new sample \(x_n\) to compute a log likelihood, which estimates the KL-divergence \(d_{KL}\), we use incoming and reference samples to estimate the MMD distance \(d_k\).

We can define the KCUSUM algorithm with the help of our previously defined MMD statistic. Given \(\delta > 0\) (\(\delta\) is detailed below), we can define \(g_\delta\) as:

\begin{equation}
    g_\delta ((x_o,x_1),(y_o,y_1)) = k(x_o,x_1) + k(y_o,y_1) -k(x_o,y_1) - k(x_1,y_o) - \delta
\end{equation}

The full step procedure of the KCUSUM algorithm is listed in Algorithm V.1. Let us assume that a change has occurred where one distribution \(p_1\), changes to \(p_2\). During even numbered iterations, the current sample \(x_n\) is paired with the previous sample \(x_{n-1}\). Then, the two points are compared with two reference samples \(y_n, y_{n-1}\) via MMD. The constant \(\delta > 0\) is subtracted from the result to get \(v_n\). The variable \(v_n\) is added to the running threshold statistic \(Z_{n-1}\) which becomes \(Z_n\). If the new value \(Z_n\) is non-positive, the algorithm restarts. During odd numbered iterations, \(Z_n\) is not adjusted. Therefore, KCUSUM can only detect changes at even numbered sample indexes.

The motivation behind subtracting \(\delta\) between each iteration of KCUSUM is to guarantee a negative pressure on the pre-change data and positive pressure in the post-change calculations. This allows one to formalize non-trivial bounds on the algorithm's behavior. If we observe the KCUSUM procedure, it can be seen that \(\mathbb{E}[v_n] = -\delta < 0\) before a change occurs, and \(\mathbb{E}[v_n] = d^2k (p_1, p_2) - \delta \) after the change. Therefore, KCUSUM can find changes on any distribution \(p_2\) that is a minimum \(\surd \delta\) away from the reference \(p_1\).

\(\textbf{Example\hspace{1mm} V.I}\)

Let us consider detecting a change in the deviation of a normal distribution, similar to the previous example for CUSUM. We have again the incoming stream of observations and our reference sample, with a change at \(t = 200\) shown as the red dashed line. The KCUSUM algorithm is based on the linear statistic \(P_L\), so at each time \(n\), the MMD estimate \(k(x_{n-1}, x_n) + k(y_{n-1}, y_n) - k(x_n, y_{n-1}) - k(x_{n-1}, y_n)\) is computed. Similar to the Cumulative Sum algorithm, a user defined threshold \(h\) decides whether or not a change exists. With the stated parameters, a threshold of \(h = 5\) with drift \(\delta = \frac{1}{40}\), found a change at \(T_{KCUSUM} = 225\).

\begin{center}
\begin{minipage}{.6\textwidth}
\begin{algorithm}[H]
\centering
    \caption{\textbf{-V.I:} Kernel CUSUM (KCUSUM)}\label{kcusumAlgo}
    \begin{algorithmic}
        \State \textbf{input:}Thresholds \(h\geq 0,\delta \ge \) and data \(x_1,x_2,\ldots\)
        \State \textbf{initialize} \(Z_1=0\)
        \State \textbf{For} \(n = 2,3,\ldots\)\textbf{do}
                \State \hspace{3mm} \textbf{sample} \(y_n\) from reference measure \(p_o\)
                \State \hspace{4mm} \textbf{if} \(n\) is even \textbf{then}
                \State \hspace{5mm} \(v_n=g_\delta ((x_o,x_1),(y_o,y_1))\)
                \State \hspace{3mm} \textbf{Else}
                \State \hspace{5mm} \(v_n\)
                \State \hspace{3mm} \textbf{end}
                \State \hspace{3mm} \(Z_n = max\{0,Z_{n-1}+v_n\}\)
                \State \hspace{3mm} \textbf{if} \(Z_n \ge h\),\textbf{then} set \(T_{KCUSUM}=n\) and \textbf{exit}
                \State \hspace{3mm} \textbf{Else} continue
                \State \textbf{end}
    \end{algorithmic}
\end{algorithm}
\end{minipage}
\end{center}

As found for the CUSUM algorithm, we now consider the rate of false alarms (ARL2FA) and average detection delay (ESADD). Using the filtration \(\{F_n\}\) where \(F_n = \sigma(x_1, y_1,\ldots, x_n, y_n)\), we can apply the previously defined Equations (1) and (2) to find ARL2FA and ESADD for the KCUSUM algorithm. To prove our upper and lower bounds, we adapt the methods of [1], which lets one reduce the problem of CUSUM analysis to the analysis of a random walk with i.i.d. terms. For convenience, we will group the variables together as \(\{Z_n\}_{n\geq 1}\) where for \(n \geq 1\):

\begin{equation}
    Z_n = (x_{2n-1},x_{2n},y_{2n-1},y_{2n})
\end{equation}

This grouping reflects how KCUSUM processes data in pairs, and as a consequence, the bounds shown below involve additional factors of two compared to our CUSUM analysis (SEE PROPOSITION II.2). We define the associated auxiliary stopping times \(c_1,\ldots ,c_n\) as:

\begin{equation}
    c_n = inf\{k\geq n|\sum_{i=n}^k g_\delta(Z_i)>h\}
\end{equation}

\(\textbf{Theorem\hspace{1mm} V.II}\)

Let \(T_{KCUSUM}\) be the change detector as described in Algorithm V.I. Let \(p_1\) be the pre-change sample distribution. If we define \(\delta > 0\), the time to false alarm is at minimum:

\begin{equation}
    ARL2FA_{KCUSUM}\geq \frac{2}{\mathbb{P}_\infty (c_1 < \infty)}
\end{equation}

If \(p_2\) is a distribution with \(d_k(p_1, p_2) > \surd \delta\), then the worst case detection delay is at most:

\begin{equation}
    ESADD_{KCUSUM}\leq 2\mathbb{E}_1[c_1]
\end{equation}

\(\textbf{Proof for\ Theorem\hspace{1mm} V.II}\)

Let \(b_o = 0\) and for \(n \geq 1\) let \(b_n = max\{0, b_{n-1} + g_\delta(Z_n)\}\). Define the stopping time \(c\) as:

\begin{equation}
    c = inf\{n\geq 1 | b_n > h\}
\end{equation}

The relation between \(c\) and the KCUSUM stopping time \(T_{KCUSUM}\) is given as:

\begin{equation}
    T_{KCUSUM} = 2c
\end{equation}

Note, as discussed in [4], c can be represented as follows:

\begin{equation}
    c = inf_{n\geq 1} c_n
\end{equation}

where each stopping time \(c_n\) uses the same decision rule, and the only difference being that they operate on shifted versions of the input sequence \(\{x_i\}i\geq 1\). In this way, Theorem 2 from [1] can be applied, which yields a lower bound of \(c\) under \(P_\infty\):

\begin{equation}
    \mathbb{E}_\infty[c]\geq \frac{1}{\mathbb{P}_\infty(c_1 < \infty)}
\end{equation}

By substituting (14) into (16), we arrive at (11). To see the upper limits of ESADD, we take the sequence \(\{x_i\}_{i\geq 1}\), which has a predetermined change on a non-even set, \(t = 2m - 1\) for \(m \geq 1\).

\begin{equation}
    Z_i ~
  \begin{cases}
    p_1\times p_1\times p_1\times p_1  & \text{for \(1\leq i < m\)} \\
    p_2\times p_2\times p_1\times p_1 & \text{for \(m\leq i\)} \\
  \end{cases}
\end{equation}

Now, here we can reason as in Theorem 2 of \cite{Lorden71}:

\begin{align} 
\begin{split}
    \mathbb{E}_{2m-1}&[(T-(2m-1))^+]|\mathcal{F}_{2(m-1)}] \\ 
    \stackrel{A}{=}&\mathbb{E}_{2m-1}[(2c-2m+1)^+|\mathcal{F}_{2(m-1)}] \\
    \stackrel{B}{\leq}&\mathbb{E}_{2m-1}[(2c_m-2m+1)^+|\mathcal{F}_{2(m-1)}] \\
    \leq&2\mathbb{E}_{2m-1}[(c_m-m+1)^+|\mathcal{F}_{2(m-1)}] \\
    \stackrel{C}{=}&2\mathbb{E}_{2m-1}[(c_m-m+1)^+] \\
    \stackrel{D}{=}&2\mathbb{E}_1[(c_1-1+1)^+] \\
    =&2\mathbb{E}_1[c_1]
\end{split}    
\end{align}

Step \textbf{A} follows from Equation (14) and Step \textbf{B} follows since \(c\) is the infimum of the \(\{c_n\}_{n\geq 1}\). Step \textbf{C} follows from the independence of \(c_m\) from \(F_{2(m-1)}\), and finally Step \textbf{D} follows from the fact that the distribution \(c_m - m\) under \(t = 2m-1\) is the same as the distribution of \(c_1 - 1\) under \(t = 1\).

An interesting case arises when a change occurs during an even sample. We consider a sequence that does not contain a rapid change, where samples are obtained from three distinct distributions:

\begin{equation}
    Z_i ~
  \begin{cases}
    p_1\times p_1\times p_1\times p_1  & \text{for \(1\leq i < m\)} \\
    p_1\times p_2\times p_1\times p_1 & \text{for \(i = m\)} \\
    p_2\times p_2\times p_1\times p_1 & \text{for \(m<i\)}
  \end{cases}
\end{equation}

By using the same reasoning in Equation (18), we have:

\begin{align} 
\begin{split}
    \mathbb{E}_{2m}&[(T-2m)^+]|\mathcal{F}_{2m-1}] \\ 
    =&\mathbb{E}_{2m}[(2c-2m)^+|\mathcal{F}_{2m-1}] \\
    \leq&2\mathbb{E}_{2m}[(c_{m+1}-m)^+|\mathcal{F}_{2m-1}] \\
    =&2\mathbb{E}_{2m}[(c_{m+1}-m)^+] \\
    =&2\mathbb{E}_2[(c_2-1)^+] \\
    =&2\mathbb{E}_1[(1+c_1-1)^+] \\
    =&2\mathbb{E}_1[c_1]
\end{split}    
\end{align}

If we combine Equations (18) and (20), we can see that for all \(t \geq 1\):

\begin{equation}
    \mathbb{E}_t[(T-t)^+|\mathcal{F}_{t-1}]\leq 2\mathbb{E}_1[c_1]
\end{equation}

By the assumption our kernel is bounded by \(||k||_{inf}\), the specific bounds as described are maintained. Since the kernel is bounded, we also assume that \(\delta\) is similarly bounded by \(2||k||_{inf}\). Therefore, if \(g_{\delta}(z)\leq 0\) for all \(z\), then the algorithm cannot detect the change.

\-\ \textbf{Corollary\hspace{1mm} V.III}

Let us assume that Theorem V.II holds, and that \(k\) and \(g_{\delta}(z)\) are bounded by the previous proof. To determine ARL2FA,

\begin{equation}
    ARL2FA_{KCUSUM}\geq 2e^{\frac{h}{4||k||_\infty}log(1+\frac{\delta}{4||k||_\infty})}
\end{equation}

If \(d_k(p_1, p_2) > \surd \delta\) describes the distribution \(p_2\), then ARL2FA has an upper bound of:

\begin{equation}
    ESADD_{KCUSUM}\leq \frac{2h}{d_k(p_1,p_2)^2-\delta}+\frac{8||k||^2_\infty}{(d_k(p_1,p_2)^2-\delta)^2}
\end{equation}

\(\textbf{Proof for\ Corollary\hspace{1mm} V.III}\)

We start with

\begin{equation}
\begin{split}
    \mathbb{P}_{\infty}(c_1 < \infty)  = \mathbb{P}_{\infty}(inf\{ k \geq 1 | \sum^k_{i=1}g_{\delta}(z_i)>h\}<\infty) \\
     =\mathbb{P}_{\infty}(sup_{k\geq 1}\sum^k_{i=1}g_{\delta}(z_i)>h)
\end{split}
\end{equation}

\begin{equation}
    \mathbb{P}_{\infty}(sup_{k\geq 1}\sum^k_{i=1}g_{\delta}(z_i)>h)\leq e^{-rh}
\end{equation}

to identify an \(r\) to satisfy our conditions. To find the set of solutions, we expand our moment generating function \(M(r)=E_{inf}[e^{rg_{\delta}(z)}]\):

\begin{equation}
    M(r)=1-r\delta + \int^r_0 \int^\lambda_0 \mathbb{E}_\infty [e^{ug_\delta(z)}g_\delta(z)^2]du d\lambda
\end{equation}

Under the assumption that \(\delta<2||k||_\infty\), it holds that \(|g_{\delta}(z)|\leq 4||k||_{\infty}\) and:

\begin{equation}
\begin{split}
    M(r)\leq 1-r\delta + 16||k||_{\infty}^2 \int^r_0 \int^\lambda_0 e^{u4||k||_{\infty}} du\ d\lambda \\
    = 1-r\delta + 16||k||_{\infty}^2 \int^r_0 \frac{1}{4||k||_{\infty}} (e^{\lambda4||k||_{\infty}}-1) d\lambda \\
    = 1-r\delta + 4||k||_{\infty} \int^r_0 (e^{\lambda4||k||_{\infty}}-1) d\lambda
\end{split}    
\end{equation}

If we minimize the with right hand side with respect to \(r\), it yields:

\begin{equation}
    r=\frac{1}{4||k||_{\infty}}log(1+\frac{\delta}{4||k||_{\infty}})
\end{equation}

The equations (11),\ (24) and (25) can be combined with the previous definition of \(r\) (28), which yields the result of (22). For the delay, note that \(\mathbb{E}_1[c_1]\) is the expectation time of a random walk crossing the upper boundary, under positive drift. Therefore, we can apply Proposition A.1, which shows:

\begin{equation}
\begin{split}
    \mathbb{E}_1[c_1]\leq \frac{h}{\mathbb{E}_1[g_{\delta}(z_1)]}+\frac{\mathbb{E}_1[(g_{\delta}(z_1)^+)^2]}{\mathbb{E}_1[g_{\delta}(z_1)]^2} \\[10pt]
    = \frac{h}{d_k(p_o,p_1)^2-\delta}+ \frac{\mathbb{E}_1[(g_{\delta}(z_1)^+)^2]}{(d_k(p_o,p_1)^2-\delta)^2}
    \end{split}
\end{equation}

By combining (12) and (29) with the bound \(g_{\delta}(z)^+\leq 2||k||_\infty\), we find the relation (23). As discussed in Theorem V.II, there is a logarithmic relationship between delay and time to false alarm. If we compute the smallest value of \(h\) that guarantees an alert rate \(x\) for each level of false alarm \(x\in\{1,2,\ldots,10^4\}\), the delay can be calculated according to (23). The analysis was performed on a theoretical problem with the parameters \(d_k(p_o,p_1)^2=\frac{1}{6}, \delta=2^-5\) and \(||k||_\infty=0.5\).
\section{Results} \label{res}
In this section, we evaluate the KCUSUM algorithm on a variety of change detection tasks. Motivated by our prior theoretical analysis and the results of \cite{Flynn19}, we apply the KCUSUM algorithm to more complex data structures. These structures allow the bounding of general performance for each type of task, and are expected use cases for the algorithm.  For each case, observations consist of vectors in \(\mathbb{R}^4\) based in Cartesian coordinates, and grouped by time. The reference data is a set of observable sampled from the pre-change distribution and is unique for each task.

\begin{enumerate}
    \item Change of mean. Atomic collisions at thermal equilibrium with one scaled component 
    \item Change of mean. 1H9T protein with partial structure deformation.
    \item Change of mean and variance. 1FME peptide with fast folding properties (\(\beta - \beta - \alpha\)).
\end{enumerate}


\begin{figure}[htbp!]

\centering
\includegraphics[width=6cm]{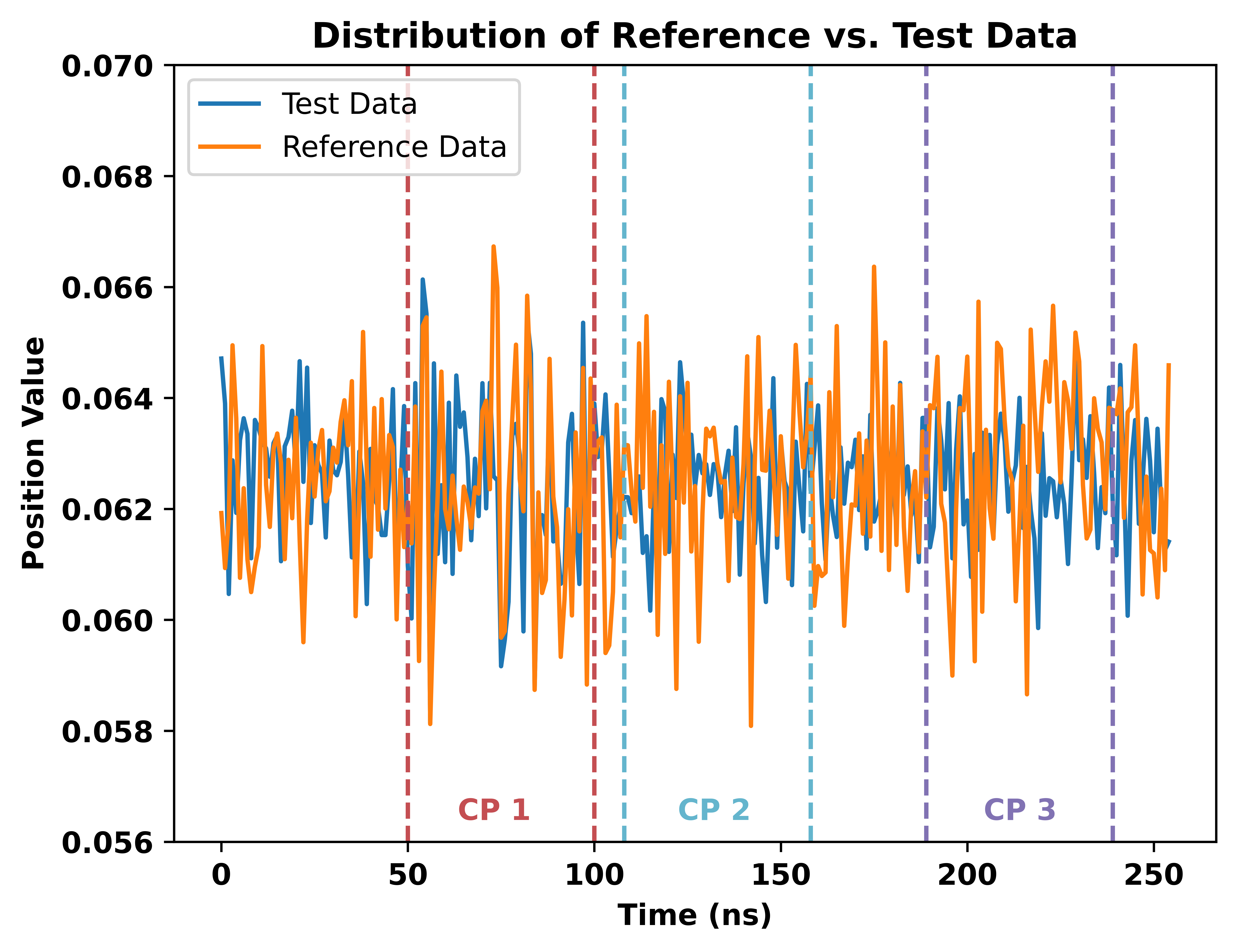}

\caption{Distribution of reference (orange) and testing data (blue) for Case 1. Each change point falls in between the set of dashed lines.}
\label{Case1_Dataplot}
\end{figure}
\vspace*{\fill} 
\begin{figure}[htbp!]\label{Case2_Sample_MMD}
\centering
\includegraphics[width=6cm]{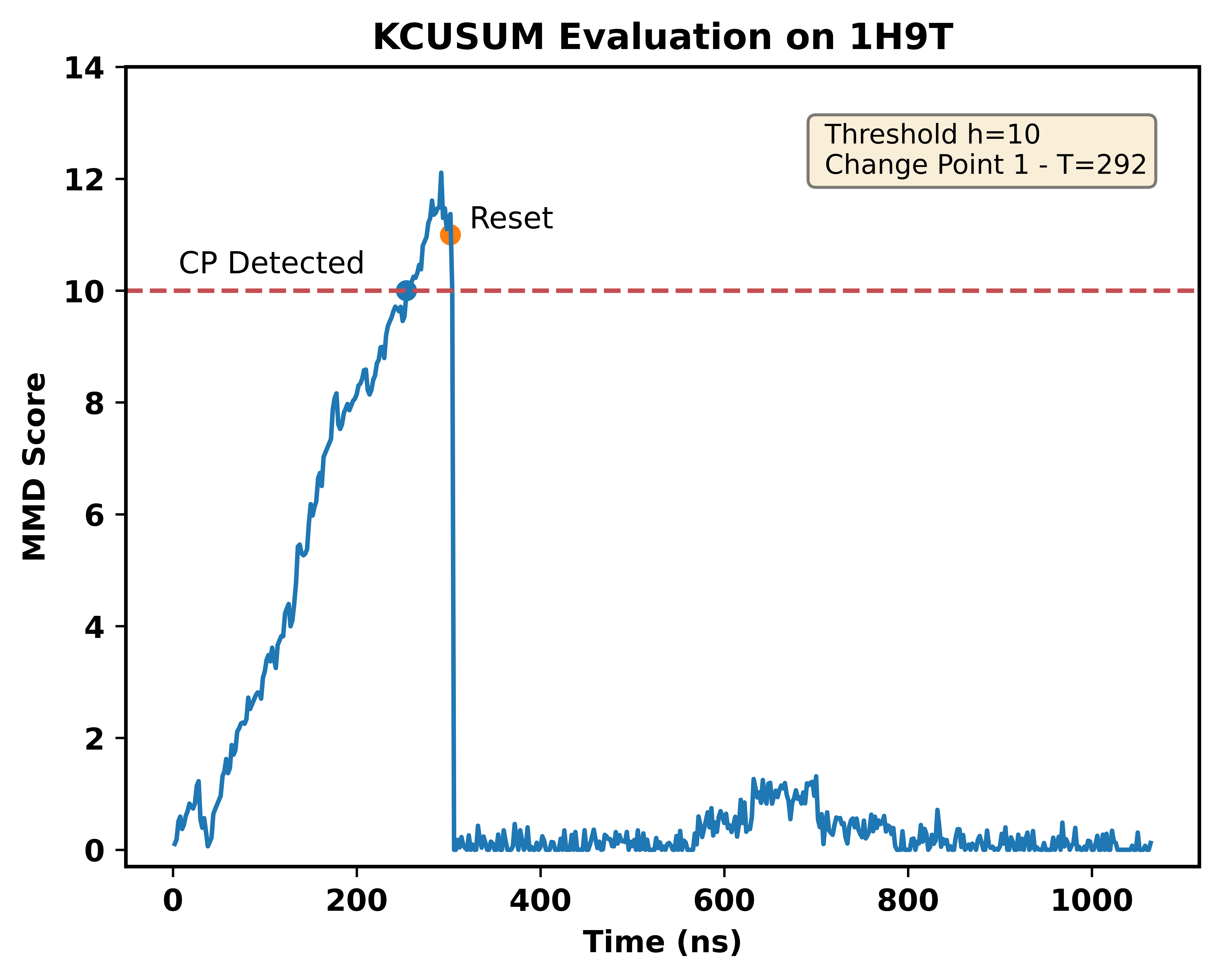}
\caption{A sample evaluation chart of the KCUSUM algorithm on data from Case 2.}
\end{figure}

\begin{figure}[htbp!]\label{1H9T_stiched}
\centering
\includegraphics[width=12cm]{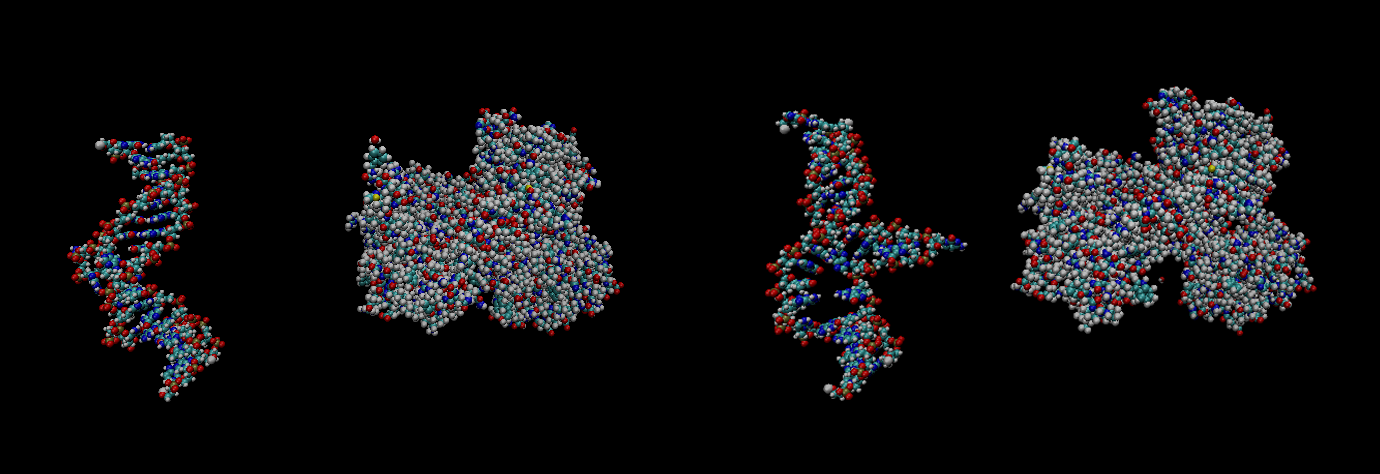}
\caption{1H9T structure before and after the change was detected.}
\end{figure}

\begin{figure}[htbp!]

\centering
\includegraphics[width=7cm,height=5cm]{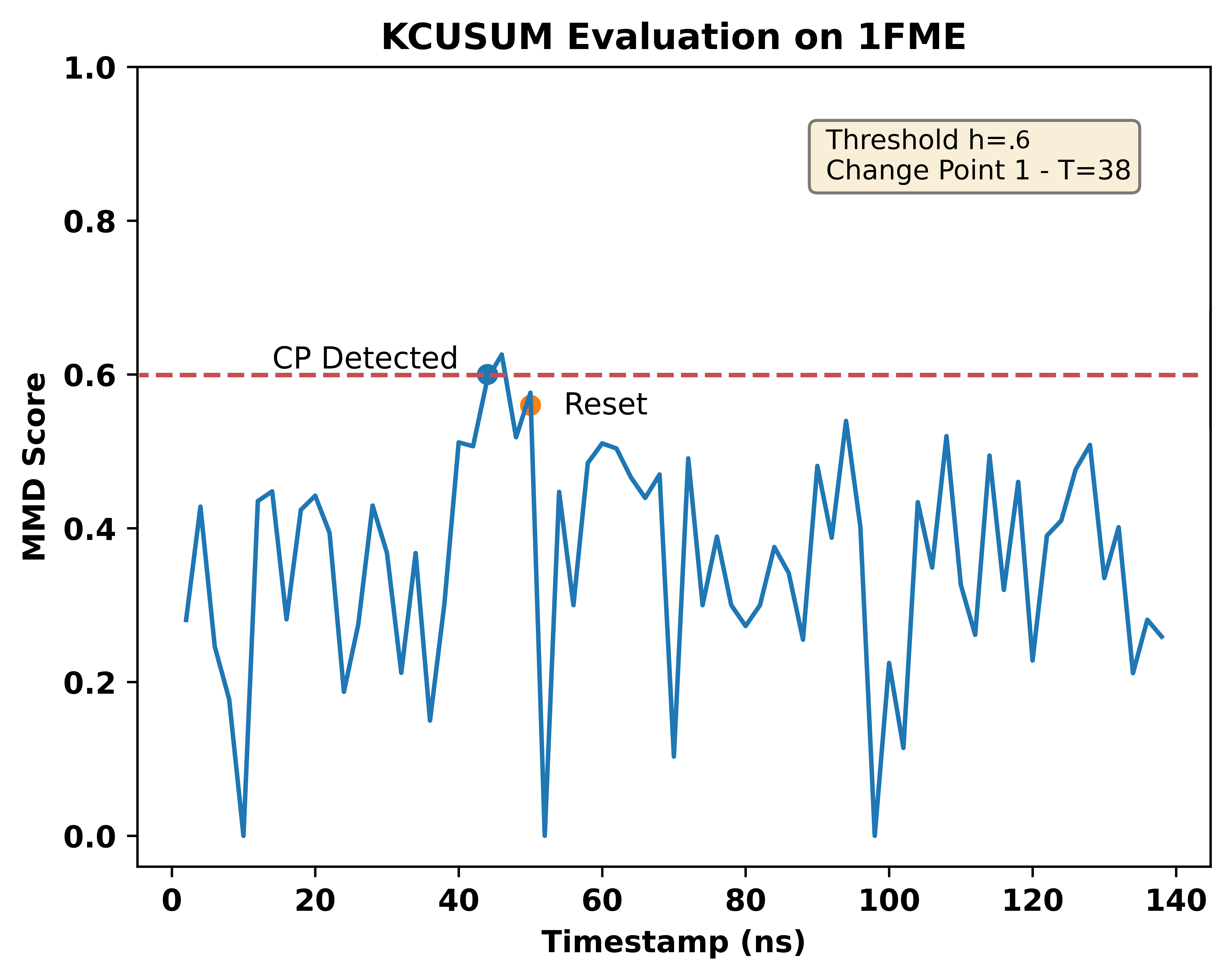}

\caption{A sample evaluation chart of the KCUSUM algorithm on data from Case 3.}
\label{Case3_Sample_MMD}
\end{figure}

\begin{figure}[htbp!]
\label{1FME_stitched}
\centering
\includegraphics[width=5cm,height=5cm]{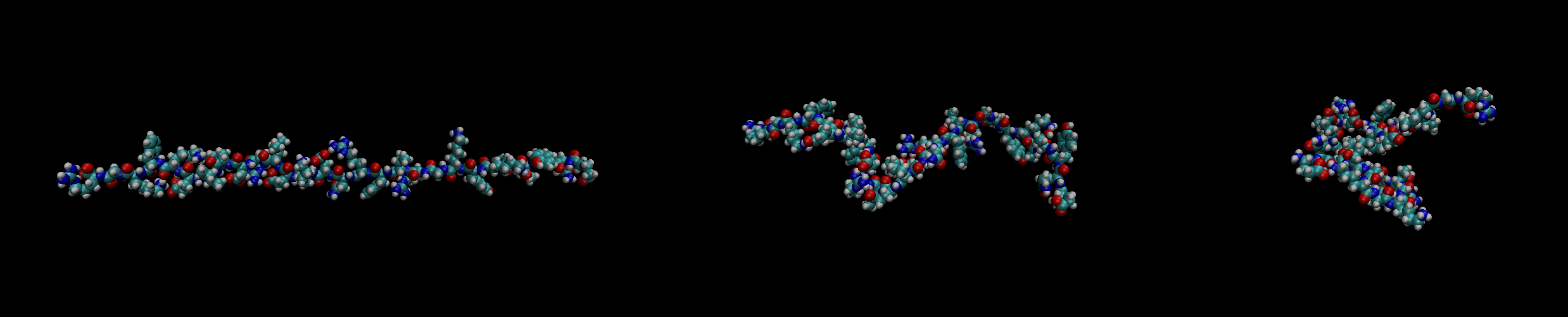}
\caption{1FME structure before, during and after the change was detected.}
\end{figure}

\begin{figure}[htbp!]
\label{Log_Plotted}
\centering
\includegraphics[width=6cm, height=6cm]{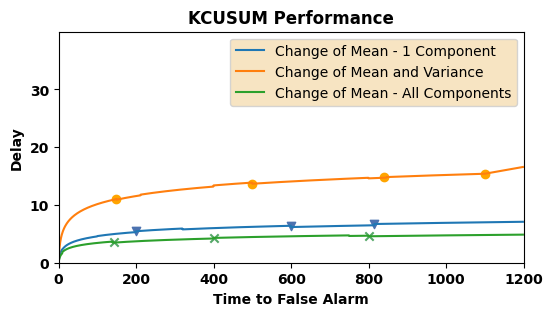}
\caption{The performance of the KCUSUM algorithm on the proposed change detection tasks.}
\end{figure}

In each case, a Monte Carlo approach is used to obtain performance bounds on KCUSUM. To test delay, each dataset is masked with a normally distributed (0,1) noise layer to generate \(n=10000\) unique sequences. The algorithm is then run and each delay time is recorded, to obtain an estimate for ESADD. The actual delay time \(t\) is independent for each task. To obtain an estimate for ARL2FA, we observe cases where a false alarm is raised before the real change point occurs, where the time is taken from the first observable.

For each case, we see a logarithmic relationship between delay and time to false alarm, which can vary in scale. This scale is an indicator of the performance of KCUSUM for each kind of task. From Figures \ref{Case1_Dataplot}-\ref{Case3_Sample_MMD}, the Case 1 and 3 change points are more easily detected, while Case 2 proves more difficult for the algorithm to detect. It is important to note for each case, that the data input to the algorithm is the incoming observables and samples drawn from the pre-change distribution; there is no information given about the post-change distribution. These results suggests that KCUSUM may be a promising change detector for problems where there is little information about the kind of change that occurs.  Additionally, when a change point is detected, the algorithm is reset so that it continues to process incoming data. When the reset occurs, a new reference sample is taken from the post-change distribution, which becomes our new reference. After a significant conformational change, the original reference sample would likely increase ARL2FA if used past the detected change point. The rate at which ARL2FA changes from reference selection is a current focus of our research. 



\section{Conclusion}
In conclusion, the evaluation of the KCUSUM algorithm on various change point detection tasks has provided valuable insights into its performance across different molecular dynamics trajectory structures. The algorithm was applied to complex data structures based on prior theoretical analysis and reference data unique to each task. Three specific cases were examined, involving changes in mean, mean and variance, and complex atomic structures.
A Monte Carlo approach was employed to establish performance bounds on KCUSUM for each case. The evaluation included testing delay through the introduction of noise layers to generate unique sequences. The algorithm's performance was assessed in terms of delay and time to false alarm, revealing a logarithmic relationship between the two, with varying scales indicative of the algorithm's effectiveness for each task.
Notably, the algorithm demonstrated effectiveness in cases involving changes in mean and atomic structure deformation, while facing challenges in detecting changes in mean and variance for a specific peptide with fast folding properties. The logarithmic relationship observed suggests that KCUSUM may be well-suited for change point detection in scenarios where little information is available about the nature of the change.
Real data evaluation was also conducted, with visual representations provided for each case. These figures illustrated the distribution of reference and testing data, as well as sample evaluation charts showcasing the algorithm's performance.
In summary, the theoretical analysis and empirical evaluations collectively suggest that KCUSUM holds promise as a change detector, particularly in situations where limited information is available about the type of change occurring. Further applications and testing in real-world scenarios will contribute to a deeper understanding of its capabilities and potential use cases.
\section*{Acknowledgements}
This material is based upon work supported by the Advanced Scientific Computing Research Program in the U.S Department of Energy, Office of Science, under FAIR (Funding for Accelerated, inclusive research) Award number DE-SC0024492. We are grateful for the support received from the SHI SRP-HPC ECP project funded by the U.S DOE, and for the excellent mentorship from the BNL scientists.

\section*{Appendix}
\textbf{Proposition II.I}:
Assume \(\{a_i\}_{i\geq 1}\) are i.i.d real-value random variables such that \(\mathbb{E}[a_i]=\mu >0\) and \(\mathbb{E}[(a_i^+)^2]<\infty\). Let us define \(S_n\) and \(T\) for \(a\leq 0 \leq b\) such that \(S-n=\sum^n_{i=1}a_i\) and \(T=inf\{n\geq 1 | S_n \notin [a,b]\}\). Now:

\begin{equation*}
    \mathbb{E}[T]\leq \frac{(1-\alpha)b+a\alpha}{\mu}+\frac{\mathbb{E}[(a_1^+)^2]}{{\mu}^2}
\end{equation*}

where \(\alpha = \mathbb{P}(S_T<a)\).

\textbf{Lemma II.II}

Assume \(\{a_i\}_{i\geq 1}\) is a sequence of i.i.d. real-value random variables such that for \(n\geq 1\), we can define the partial sum as \(S_n=\sum^n_{i=1}a_i\). Let \(M(r)=\mathbb{E}[e^{ra_1}]\) be our moment generating function for \(a_i\) and assume for some \(q>0\), there is a \(M(q)\leq 1\). Now, for any \(h\geq 0\), we have:

\begin{equation*}
    \mathbb{P}(sup_{n\geq 1}S_n>h)\leq e^{-qh}
\end{equation*}

\-\ \(Proof of Lemma II.II:\)
\newline
Let us define \(Z-n=e^{qS_n}\) for all \(n\geq 1\). Now, we can show for \(n\geq 1\):

\begin{equation*}
\begin{split}
    \mathbb{E}[Z_{n+1}|Z_n]=\mathbb{E}[\prod^{n+1}_{i=1}e^{qa_i}|Z_n] \\
    =Z_n\mathbb{E}[e^{qa_{n+1}}] \\
    =Z_nM(q)\leq Z_n
    \end{split}
\end{equation*}

Then, for all such \(n\geq 1\), it holds that \(Z_n\geq 0\) and \(\mathbb{E}[|E|]=M(q)^n\leq 1\). Since \(Z_n\) is a non-negative supermartingale, then we have:

\begin{equation*}
\begin{split}
    \mathbb{P}(sup_{n\geq 1}S_n>h)=\mathbb{P}(sup_{n\geq 1}Z_n>e^{-qh}) \\
    \leq \mathbb{E}[e^{qa_1}]e^{-qh} \\
    \leq e^{-qh}
\end{split}
\end{equation*}

The first part in the previous derivation follows from the monotonic function \(x\rightharpoonup e^x\), where the second part is a result of Theorem 7.8 \cite{Gallager95}. The final line is a result of our assumption on \(q\).

\-\ \(Proof of Proposition\hspace{1mm} II.I\):

Let us define \(S_1,S_2,\ldots\) as:

\begin{equation*}
    S_n=\sum^n_{i=1}log\frac{f_1(x_i)}{f_o(x_i)}
\end{equation*}

with \(\alpha =\mathbb{P}_\infty(sup_{n\geq 1}S_n>h)\). Theorem 2 \cite{Lorden71} shows that CUSUM follows:

\begin{equation*}
    ARL2FA_{CUSUM}\geq \frac{1}{\alpha}
\end{equation*}

Since \(\mathbb{E}_\infty[log\frac{f_1(x_i)}{f_o(x_i)}]=-d_{KL}(p_o,p_1)<0\), \(S_n\) is a random walk with a negative drift. The moment generating function can be shown as:

\begin{equation*}
M(r)=\mathbb{E}[e^{rlog\frac{f_1(x_i)}{f_o(x_i)}}]=\mathbb{E}_\infty[(\frac{f_1(x_i)}{f_o(x_i)})^r]
\end{equation*}

From \(M(r)\), we can see that \(g(1)=\mathbb{E}_1=1\). Now we can apply Lemma II.II with \(q=1\) to show that \(\alpha \leq e^{-h}\).

Now let \(N\) be our stopping time \(N=inf\{n\geq 1|S_n>h\}\). If we apply Theorem 2 \cite{Lorden71}, it shows that:

\begin{equation*}
    ESADD_{CUSUM}\leq \mathbb{E}_1[N]
\end{equation*}

Assume that a change occurs at \(t=1\), and that the variables \(\frac{f_1(x_i)}{f_o(x_i)};i=1,2,\ldots\) are i.i.d. with a positive mean. Since \(\{S_n\}_{n\geq 1}\) is a random walk with a positive drift \(\mu = \mathbb{E}_1[log\frac{f_1(x_i)}{f_o(x_i)}]\), it follows from Proposition II.I with stopping time \(N\) on the established bounds of ESADD.

\bibliographystyle{ACM-Reference-Format}
\bibliography{refs}

\end{document}